\documentstyle[aclap,epsf]{article} 

\title{\vspace{-0.5in}Document Classification Using a Finite
Mixture Model\\
}
\author{Hang Li \ \ \ \ Kenji Yamanishi\\
{C\&C Res. Labs., NEC}\\
4-1-1 Miyazaki Miyamae-ku Kawasaki, 216, Japan\\
Email: \{lihang,yamanisi\}@sbl.cl.nec.co.jp\\}

\begin{document}

\maketitle
\vspace{-0.5in}
\begin{abstract}

  We propose a new method of classifying documents into categories. We
  define for each category a {\em finite mixture model} based on {\em
    soft clustering} of words. We treat the problem of classifying
  documents as that of conducting statistical hypothesis testing over
  finite mixture models, and employ the EM algorithm to efficiently
  estimate parameters in a finite mixture model.  Experimental results
  indicate that our method outperforms existing methods.

\end{abstract}

\section{Introduction} 

We are concerned here with the issue of classifying documents into
categories. More precisely, we begin with a number of categories
(e.g., `tennis, soccer, skiing'), each already containing certain
documents. Our goal is to determine into which categories newly given
documents ought to be assigned, and to do so on the basis of the
distribution of each document's words.\footnote{A related issue is the
  retrieval, from a data base, of documents which are relevant to a
  given query (pseudo-document)
  (e.g.,\cite{Deerwester90,Fuhr89a,Robertson76,Salton83,Wong89}).}

Many methods have been proposed to address this issue, and a number of
them have proved to be quite effective
(e.g.,\cite{Apte94,Cohen96,Lewis92a,Lewis94b,Lewis96,Schutze95,Yang94}).
The simple method of conducting hypothesis testing over word-based
distributions in categories (defined in Section 2) is not efficient in
storage and suffers from the {\em data sparseness problem}, i.e., the
number of parameters in the distributions is large and the data size
is not sufficiently large for accurately estimating them. In order to
address this difficulty, \cite{Guthrie94} have proposed using
distributions based on what we refer to as {\em hard clustering} of
words, i.e., in which a word is assigned to a single cluster and words
in the same cluster are treated uniformly.  The use of hard clustering
might, however, degrade classification results, since the
distributions it employs are not always precise enough for
representing the differences between categories.

We propose here to employ {\em soft clustering}\footnote{We borrow
  from \cite{Pereira93} the terms hard clustering and soft clustering,
  which were used there in a different task.}, i.e., a word can be
assigned to several different clusters and each cluster is
characterized by a specific word probability distribution. We define
for each category a {\em finite mixture model}, which is a linear
combination of the word probability distributions of the clusters. We
thereby treat the problem of classifying documents as that of
conducting statistical hypothesis testing over finite mixture models.
In order to accomplish hypothesis testing, we employ the EM algorithm
to efficiently and approximately calculate from training data the
maximum likelihood estimates of parameters in a finite mixture model.

Our method overcomes the major drawbacks of the method using
word-based distributions and the method based on hard clustering,
while retaining their merits; it in fact includes those two methods as
special cases. Experimental results indicate that our method
outperforms them.

Although the finite mixture model has already been used elsewhere in
natural language processing (e.g. \cite{Jelinek80,Pereira93}), this is
the first work, to the best of knowledge, that uses it in the context
of document classification.

\section{Previous Work}
\subsection*{Word-based method}
A simple approach to document classification is to view this problem
as that of conducting hypothesis testing over word-based
distributions. In this paper, we refer to this approach as the {\em
  word-based method} (hereafter, referred to as WBM).

Letting $W$ denote a vocabulary (a set of words), and $w$ denote a
random variable representing any word in it, for each category $c_i$
$(i=1,\cdots,n)$, we define its {\em word-based distribution}
$P(w|c_i)$ as a histogram type of distribution over $W$. (The number
of free parameters of such a distribution is thus $|W|-1$). WBM then
views a document as a sequence of words:
\begin{equation} d= w_{1},\cdots,w_{N}
\end{equation}
and assumes that each word is generated independently according to a
probability distribution of a category. It then calculates the
probability of a document with respect to a category as
\begin{equation}\label{eq:wbmlike} P(d|c_i) =
  P(w_{1},\cdots,w_{N}|c_i)=\prod_{t=1}^{N}P(w_{t}|c_i),
\end{equation} 
and classifies the document into that category for which the
calculated probability is the largest. We should note here that a
document's probability with respect to each category is equivalent to
the {\em likelihood} of each category with respect to the document,
and to classify the document into the category for which it has the
largest probability is equivalent to classifying it into the category
having the largest likelihood with respect to it. Hereafter, we will
use only the term likelihood and denote it as $L(d|c_i)$.

Notice that in practice the parameters in a distribution must be
estimated from training data. In the case of WBM, the number of
parameters is large; the training data size, however, is usually not
sufficiently large for accurately estimating them. This is the {\em
  data sparseness problem} that so often stands in the way of reliable
statistical language processing (e.g.\cite{Gale90}).  Moreover, the
number of parameters in word-based distributions is too large to be
efficiently stored.

\subsection*{Method based on hard clustering}
In order to address the above difficulty, Guthrie et.al. have proposed
a method based on hard clustering of words \cite{Guthrie94} (hereafter
we will refer to this method as HCM). Let $c_1,\cdots,c_n$ be
categories.  HCM first conducts hard clustering of words.
Specifically, it (a) defines a vocabulary as a set of words $W$ and
defines as clusters its subsets $k_1,\cdots,k_m$ satisfying
$\cup_{j=1}^{m} k_j = W$ and $k_i \cap k_j = \emptyset$ $(i\not= j)$
(i.e., each word is assigned only to a single cluster); and (b) treats
uniformly all the words assigned to the same cluster. HCM then defines
for each category $c_i$ a distribution of the clusters $P(k_j|c_i)$
$(j=1,\cdots,m)$. It replaces each word $w_{t}$ in the document with
the cluster $k_{t}$ to which it belongs $(t=1,\cdots,N)$. It assumes
that a cluster $k_{t}$ is distributed according to $P(k_j|c_i)$ and
calculates the likelihood of each category $c_i$ with respect to the
document by
\begin{equation}\label{eq:hcmlike} 
L(d|c_i) =
L(k_{1},\cdots,k_{N}|c_i)=\prod_{t=1}^{N}P(k_{t}|c_i).
\end{equation} 

\begin{table}[htb]
\begin{center}
\caption{Frequencies of words}
\label{tab:train}
\begin{tabular}{|l|cccccc|} \hline
 & racket & stroke & shot & goal & kick & ball \\ \hline
$c_1$ & $4$ & $1$ & $2$ & $1$ & $0$ & $2$ \\
$c_2$ & $0$ & $0$ & $0$ & $3$ & $2$ & $2$ \\ \hline
\end{tabular}
\end{center}
\end{table}

\begin{table}[htb]
\begin{center}
\caption{Clusters and words ($L=5$,$M=5$)}
\label{tab:hardclus}
\begin{tabular}{|l|l|} \hline
$k_1$ & racket, stroke, shot \\
$k_2$ & kick \\
$k_3$ & goal, ball \\ \hline
\end{tabular}
\end{center}
\end{table}

\begin{table}[htb]
\begin{center}
\caption{Frequencies of clusters}
\label{tab:hardfreq}
\begin{tabular}{|l|ccc|} \hline
 & $k_1$ & $k_2$ & $k_3$ \\ \hline
$c_1$ & $7$ & $0$ & $3$ \\
$c_2$ & $0$ & $2$ & $5$ \\ \hline
\end{tabular}
\end{center}
\end{table}

\begin{table}[htb]
\begin{center}
\caption{Probability distributions of clusters}
\label{tab:hardprob}
\begin{tabular}{|l|ccc|} \hline
 & $k_1$ & $k_2$ & $k_3$ \\ \hline
$c_1$ & $0.65$ & $0.04$ & $0.30$ \\
$c_2$ & $0.06$ & $0.29$ & $0.65$ \\ \hline
\end{tabular}
\end{center}
\end{table}

There are any number of ways to create clusters in hard clustering,
but the method employed is crucial to the accuracy of document
classification. Guthrie et. al. have devised a way suitable to
documentation classification. Suppose that there are two categories
$c_1$=`tennis' and $c_2$=`soccer,' and we obtain from the training
data (previously classified documents) the frequencies of words in
each category, such as those in Tab.~\ref{tab:train}. Letting $L$ and
$M$ be given positive integers, HCM creates three clusters: $k_1$,
$k_2$ and $k_3$, in which $k_1$ contains those words which are among
the $L$ most frequent words in $c_1$, and not among the $M$ most
frequent in $c_2$; $k_2$ contains those words which are among the $L$
most frequent words in $c_2$, and not among the $M$ most frequent in
$c_1$; and $k_3$ contains all remaining words (see
Tab.~\ref{tab:hardclus}).  HCM then counts the frequencies of clusters
in each category (see Tab.~\ref{tab:hardfreq}) and estimates the
probabilities of clusters being in each category (see
Tab.~\ref{tab:hardprob}).\footnote{We calculate the probabilities here
  by using the so-called expected likelihood estimator
  \cite{Gale90}:\begin{equation} P(k_j|c_i) =
  \frac{f(k_j|c_i)+0.5}{f(c_i)+0.5\times m},
\end{equation} where $f(k_j|c_i)$ is the frequency of the cluster
$k_j$ in $c_i$, $f(c_i)$ is the total frequency of clusters in $c_i$,
and $m$ is the total number of clusters.} Suppose that a newly given
document, like $d$ in Fig.~\ref{fig:doc}, is to be classified. HCM
calculates the likelihood values $L(d|c_1)$ and $L(d|c_2)$ according
to Eq.~(\ref{eq:hcmlike}).  (Tab.~\ref{tab:hcmlike} shows the
logarithms of the resulting likelihood values.) It then classifies $d$
into $c_2$, as $\log_2 L(d|c_2)$ is larger than $\log_2 L(d|c_1)$.

\begin{figure}[h]
\begin{center}
$d$ = kick, goal, goal, ball
\caption{Example document}
\label{fig:doc}
\end{center}
\end{figure}

\begin{table}[htb]
\begin{center}
\caption{Calculating log likelihood values}
\label{tab:hcmlike}
\begin{tabular}{|l|} \hline
$\log_2 L(d|\mbox{$c_1$})$ \\
$= 1 \times \log_2 .04 + 3 \times \log_2 .30 $ = $-9.85$ \\ \hline
$\log_2 L(d|\mbox{$c_2$})$ \\
$= 1 \times \log_2 .29 + 3 \times \log_2 .65 $ = $-3.65$ \\ \hline
\end{tabular}
\end{center}
\end{table}

HCM can handle the data sparseness problem quite well. By assigning
words to clusters, it can drastically reduce the number of parameters
to be estimated. It can also save space for storing knowledge. We
argue, however, that the use of hard clustering still has the
following two problems:

\begin{enumerate} \item {\em HCM cannot assign a word to more than one
    cluster at a time}. Suppose that there is another category $c_3$ =
  `skiing' in which the word `ball' does not appear, i.e., `ball' will
  be indicative of both $c_1$ and $c_2$, but not $c_3$. If we could
  assign `ball' to both $k_1$ and $k_2$, the likelihood value for
  classifying a document containing that word to $c_1$ or $c_2$ would
  become larger, and that for classifying it into $c_3$ would become
  smaller. HCM, however, cannot do that.  \item {\em HCM cannot make
    the best use of information about the differences among the
    frequencies of words assigned to an individual cluster}. For
  example, it treats `racket' and `shot' uniformly because they are
  assigned to the same cluster $k_1$ (see Tab.~\ref{tab:hcmlike}).
  `Racket' may, however, be more indicative of $c_1$ than `shot,'
  because it appears more frequently in $c_1$ than `shot.' HCM fails
  to utilize this information. This problem will become more serious
  when the values $L$ and $M$ in word clustering are large, which
  renders the clustering itself relatively meaningless.
\end{enumerate}

From the perspective of number of parameters, HCM employs models
having very few parameters, and thus may not sometimes represent much
useful information for classification.

\section{Finite Mixture Model}

We propose a method of document classification based on soft
clustering of words. Let $c_1,\cdots,c_n$ be categories. We first
conduct the soft clustering. Specifically, we (a) define a vocabulary
as a set $W$ of words and define as clusters a number of its subsets
$k_1,\cdots,k_m$ satisfying $\cup_{j=1}^{m}k_j=W$; (notice that $k_i
\cap k_j = \emptyset$ $(i\not=j)$ does not necessarily hold here,
i.e., a word can be assigned to several different clusters); and (b)
define for each cluster $k_j$ $(j=1,\cdots,m)$ a distribution
$Q(w|k_j)$ over its words $(\sum_{w\in k_j}Q(w|k_j)=1)$ and a
distribution $P(w|k_j)$ satisfying: \begin{equation} P(w|k_j)=\left\{
\begin{array}{ll} Q(w|k_j); & w \in k_j, \\ 0; & w \notin k_j, \\
\end{array} \right.  \end{equation} where $w$ denotes a random
variable representing any word in the vocabulary. We then define for
each category $c_i$ $(i=1,\cdots,n)$ a distribution of the clusters
$P(k_j|c_i)$ , and define for each category a linear combination of
$P(w|k_j)$:
\begin{equation}\label{eq:fmm} P(w|c_i) = \sum_{j=1}^{m} P(k_j|c_i)
  \times P(w|k_j) \end{equation} as the distribution over its words,
  which is referred to as a {\em finite mixture model}
  (e.g., \cite{Everitt81}).

  We treat the problem of classifying a document as that of conducting
  the likelihood ratio test over finite mixture models. That is, we
  view a document as a sequence of words, \begin{equation}
  d=w_{1},\cdots,w_{N} \end{equation} where $w_{t}$$(t=1,\cdots,N)$
  represents a word. We assume that each word is independently
  generated according to an unknown probability distribution and
  determine which of the finite mixture models $P(w|c_i)
  (i=1,\cdots,n)$ is more likely to be the probability distribution by
  observing the sequence of words. Specifically, we calculate the
  likelihood value for each category with respect to the document by:
\begin{equation}\label{eq:fmmlike} \begin{array}{rcl} L(d|c_i) & = &
L(w_{1},\cdots,w_{N}|c_i) \\
 & = & \prod_{t=1}^{N} P(w_{t}|c_i) \\
  & = & \prod_{t=1}^{N} \left(\sum_{j=1}^m 
  P(k_j|c_i) \times P(w_{t}|k_j) \right). \\ \end{array}
\end{equation} We then classify it into the category having the
largest likelihood value with respect to it. Hereafter, we will
refer to this method as FMM.

FMM includes WBM and HCM as its special cases. If we consider the
specific case (1) in which a word is assigned to a single cluster and
$P(w|k_j)$ is given by
\begin{equation} P(w|k_j) = \left\{ \begin{array}{ll} \frac{1}{|k_j|};
  & w \in k_j, \\ 0; & w \notin k_j, \\ \end{array} \right.
\end{equation} where $|k_j|$ denotes the number of elements belonging
to $k_j$, then we will get the same classification result as in HCM.
In such a case, the likelihood value for each category $c_i$ becomes:
\begin{equation}\label{eq:fmmlike2}
\begin{array}{rcl} L(d|c_i)
  & = & \prod_{t=1}^{N} \left(P(k_{t}|c_i)\times P(w_{t}|k_{t})\right)
  \\ & = & \prod_{t=1}^{N} P(k_{t}|c_i) \times \prod_{t=1}^{N}
  P(w_{t}|k_{t}), \end{array}\end{equation} where $k_{t}$ is the
  cluster corresponding to $w_{t}$.  Since the probability
  $P(w_{t}|k_{t})$ does not depend on categories, we can ignore the
  second term $\prod_{t=1}^{N} P(w_t|k_t)$ in hypothesis testing, and
  thus our method essentially becomes equivalent to HCM (c.f.
  Eq.~(\ref{eq:hcmlike})).

Further, in the specific case (2) in which $m=n$, for each
$j$, $P(w|k_j)$ has $|W|$ parameters: $P(w|k_j)=P(w|c_j)$, and
$P(k_j|c_i)$ is given by
\begin{equation}\label{eq:peak}
  P(k_j|c_i) = \left\{ \begin{array}{ll} 1; & i=j, \\ 0;
  & i \not= j, \\ \end{array} \right.
\end{equation}
the likelihood used in hypothesis testing becomes the same as that in
Eq.(\ref{eq:wbmlike}), and thus our method becomes equivalent to WBM.

\section{Estimation and Hypothesis Testing}

In this section, we describe how to implement our method.

\subsection*{Creating clusters}
There are any number of ways to create clusters on a given set of
words. As in the case of hard clustering, the way that clusters are
created is crucial to the reliability of document classification.
Here we give one example approach to cluster creation.

\begin{table}[htb]
\begin{center}
\caption{Clusters and words}
\label{tab:softclus}
\begin{tabular}{|l|l|} \hline
$k_1$ & racket, stroke, shot, ball \\
$k_2$ & kick, goal, ball \\ \hline
\end{tabular}
\end{center}
\end{table}

We let the number of clusters equal that of categories (i.e., $m=n$)
\footnote{One can certainly assume that $m \ge n$.} and relate each
cluster $k_i$ to one category $c_i$ $(i=1,\cdots,n)$. We then assign
individual words to those clusters in whose related categories they
most frequently appear. Letting $\gamma$ $(0\le
\gamma <1)$ be a predetermined threshold value, if the following
inequality holds:
\begin{equation}\label{eq:gamma} \frac{f(w|c_i)}{f(w)} > \gamma,
\end{equation}
then we assign $w$ to $k_i$, the cluster related to $c_i$, where
$f(w|c_i)$ denotes the frequency of the word $w$ in category $c_i$,
and $f(w)$ denotes the total frequency of $w$. Using the data in
Tab.\ref{tab:train}, we create two clusters: $k_1$ and $k_2$, and
relate them to $c_1$ and $c_2$, respectively. For example, when
$\gamma=0.4$, we assign `goal' to $k_2$ only, as the relative
frequency of `goal' in $c_2$ is $0.75$ and that in $c_1$ is only
$0.25$. We ignore in document classification those words which cannot
be assigned to any cluster using this method, because they are not
indicative of any specific category. (For example, when $\gamma \ge
0.5$ `ball' will not be assigned into any cluster.) This helps to make
classification efficient and accurate.  Tab.~\ref{tab:softclus} shows
the results of creating clusters.

\subsection*{Estimating $P(w|k_j)$}
We then consider the frequency of a word in a cluster. If a word is
assigned only to one cluster, we view its total frequency as its
frequency within that cluster. For example, because `goal' is assigned
only to $k_2$, we use as its frequency within that cluster the total
count of its occurrence in all categories. If a word is assigned to
several different clusters, we distribute its total frequency among
those clusters in proportion to the frequency with which the word
appears in each of their respective related categories. For example,
because `ball' is assigned to both $k_1$ and $k_2$, we distribute its
total frequency among the two clusters in proportion to the frequency
with which `ball' appears in $c_1$ and $c_2$, respectively. After
that, we obtain the frequencies of words in each cluster as shown in
Tab.~\ref{tab:softclusfreq}.

\begin{table}[htb]
\begin{center}
\caption{Distributed frequencies of words}
\label{tab:softclusfreq}
\begin{tabular}{|l|cccccc|} \hline
 & racket & stroke & shot & goal & kick & ball \\ \hline $k_1$ & $4$ & $1$
& $2$ & \underline{$0$} & $0$ & \underline{$2$} \\ $k_2$ & $0$ & $0$ & $0$ &
\underline{$4$} & $2$ & \underline{$2$} \\ \hline \end{tabular} \end{center} \end{table}

We then estimate the probabilities of words in each cluster, obtaining
the results in Tab.~\ref{tab:softclusprob1}.\footnote{We calculate the
probabilities by employing the maximum likelihood estimator:
\begin{equation} P(w|k_j) = \frac{f(w|k_j)}{f(k_j)},
\end{equation} where $f(w|k_j)$ is the frequency of $w$
in $k_j$, and $f(k_j)$ is the total frequency of words in $k_j$.}

\begin{table}[htb]
\begin{center}
\caption{Probability distributions of words}
\label{tab:softclusprob1}
\begin{tabular}{|l|cccccc|} \hline
 & racket & stroke & shot & goal & kick & ball \\ \hline
$k_1$ & $0.44$ & $0.11$ & $0.22$ & $0$ & $0$ & $0.22$ \\
$k_2$ & $0$ & $0$ & $0$ & $0.50$ & $0.25$ & $0.25$ \\ \hline
\end{tabular}
\end{center}
\end{table}

\begin{table}[htb]
\begin{center}
\caption{Probability distributions of clusters}
\label{tab:softclusprob2}
\begin{tabular}{|l|cc|} \hline
 & $k_1$ & $k_2$ \\ \hline
$c_1$ & $0.86$ & $0.14$ \\
$c_2$ & $0.04$ & $0.96$ \\ \hline
\end{tabular}
\end{center}
\end{table}

\begin{table*}[htb]
\begin{center}
\caption{Calculating log likelihood values}
\label{tab:fmmlike}
\begin{tabular}{|l|} \hline
$\log_2 L(d|\mbox{$c_1$})$ = $\log_2 (.14 \times .25) + 2 \times \log_2 (.14 \times .50)+\log_2 (.86\times .22 +.14\times .25) $ = $-14.67$ \\ \hline
$\log_2 L(d|\mbox{$c_2$})$ = $\log_2 (.96 \times .25) + 2
\times \log_2 (.96 \times .50)+\log_2 (.04\times .22+.96\times .25)$
= $-6.18$ \\ \hline
\end{tabular}
\end{center}
\end{table*}

\subsection*{Estimating $P(k_j|c_i)$}
Let us next consider the estimation of $P(k_{j}|c_{i}).$ There are two
common methods for statistical estimation, the maximum likelihood
estimation method and the Bayes estimation method. In their
implementation for estimating $P(k_{j}|c_{i})$, however, both of them
suffer from computational intractability. The EM algorithm
\cite{Dempster77} can be used to efficiently approximate the maximum
likelihood estimator of $P(k_{j}|c_{i})$. We employ here an extended
version of the EM algorithm \cite{Helmbold95}. (We have also devised,
on the basis of the Markov chain Monte Carlo (MCMC) technique (e.g. 
\cite{Tanner87,Yamanishi96})\footnote{We have confirmed in our
  preliminary experiment that MCMC performs slightly better than EM in
  document classification, but we omit the details here due to space
  limitations.}, an algorithm to efficiently approximate the Bayes
estimator of $P(k_{j}|c_{i})$.)

For the sake of notational simplicity, for a fixed $i,$ let us write
$P(k_{j}|c_{i})$ as $\theta_{j}$ and $P(w|k_{j})$ as $P_{j}(w)$. Then
letting $\theta =(\theta_{1},\cdots,\theta_{m}),$ the finite mixture
model in Eq.~(\ref{eq:fmm}) may be written as
\begin{equation}
  P(w|\theta)=\sum ^{m}_{j=1}\theta_{j}\times P_{j}(w).
\end{equation} 
For a given training sequence $w_{1}\cdots w_{N},$ the maximum
likelihood estimator of $\theta$ is defined as the value
$\hat{\theta}$ which maximizes the following log likelihood function
\begin{equation}
L(\theta) = \frac{1}{N} \sum_{t=1}^{N} \log \left(\sum_{j=1}^{m}\theta_j P_j(w_t) \right).
\end{equation}

The EM algorithm first arbitrarily sets the initial value of $\theta$,
which we denote as $\theta^{(0)}$, and then successively calculates
the values of $\theta$ on the basis of its most recent values. Let $s$
be a predetermined number. At the $l$th iteration ($l=1,\cdots,s$), we
calculate $\theta^{(l)}=(\theta^{(l)}_{1},\cdots,\theta^{(l)}_m)$ by
\begin{equation}
\theta^{(l)}_{j} = \theta^{(l-1)}_j \left(\eta (\bigtriangledown
L(\theta^{(l-1)})_j-1) + 1\right),
\end{equation}
where $\eta > 0$ (when $\eta=1$, Hembold et al. 's version simply
becomes the standard EM algorithm), and $\bigtriangledown L(\theta)$
denotes
\begin{equation}
  \bigtriangledown L(\theta) = \left( \frac{\partial L}{\partial \theta_1} \cdots \frac{\partial L}{\partial \theta_m} \right).
\end{equation}
After $s$ numbers of calculations, the EM algorithm outputs
$\theta^{(s)}=(\theta^{(s)}_{1},\cdots,\theta^{(s)}_m)$ as an
approximate of $\hat{\theta}$. It is theoretically guaranteed that the
EM algorithm converges to a local maximum of the given likelihood
\cite{Dempster77}.

For the example in Tab.~\ref{tab:train}, we obtain the results as
shown in Tab.~\ref{tab:softclusprob2}.

\subsection*{Testing}
For the example in Tab.~\ref{tab:train}, we can calculate according to
Eq.~(\ref{eq:fmmlike}) the likelihood values of the two categories
with respect to the document in Fig.~\ref{fig:doc}
(Tab.~\ref{tab:fmmlike} shows the logarithms of the likelihood
values). We then classify the document into category $c_2$, as $\log_2
L(d|c_2)$ is larger than $\log_2 L(d|c_1)$.

\section{Advantages of FMM}

For a probabilistic approach to document classification, the most
important thing is to determine what kind of probability model
(distribution) to employ as a representation of a category. It must
(1) appropriately represent a category, as well as (2) have a proper
preciseness in terms of number of parameters. The goodness and badness
of selection of a model directly affects classification results.

The finite mixture model we propose is particularly well-suited to the
representation of a category. Described in linguistic terms, a cluster
corresponds to a {\em topic} and the words assigned to it are related
to that topic.  Though documents generally concentrate on a single
topic, they may sometimes refer for a time to others, and while a
document is discussing any one topic, it will naturally tend to use
words strongly related to that topic. A document in the category of
`tennis' is more likely to discuss the topic of `tennis,' i.e., to use
words strongly related to `tennis,' but it may sometimes briefly shift
to the topic of `soccer,' i.e., use words strongly related to
`soccer.' A human can follow the sequence of words in such a document,
associate them with related topics, and use the distributions of
topics to classify the document. Thus the use of the finite mixture
model can be considered as a stochastic implementation of this
process.

\begin{table}[htb] \begin{center} \caption{Num. of parameters}
\label{tab:para} \begin{tabular}{|l|c|}
\hline WBM & $O(n \cdot |W|)$ \\ HCM & $O(n \cdot
m)$ \\ FMM & $O(|k| + n \cdot m)$ \\ \hline \end{tabular}
\end{center} \end{table}

The use of FMM is also appropriate from the viewpoint of number of
parameters. Tab.~\ref{tab:para} shows the numbers of parameters in our
method (FMM), HCM, and WBM, where $|W|$ is the size of a vocabulary,
$|k|$ is the sum of the sizes of word clusters
(i.e.,$|k|=\sum_{j=1}^{m}|k_j|$), $n$ is the number of categories, and
$m$ is the number of clusters. The number of parameters in FMM is much
smaller than that in WBM, which depends on $|W|$, a very large number
in practice (notice that $|k|$ is always smaller than $|W|$ when we
employ the clustering method (with $\gamma \ge 0.5$) described in
Section 4. As a result, FMM requires less data for parameter
estimation than WBM and thus can handle the data sparseness problem
quite well. Furthermore, it can economize on the space necessary for
storing knowledge. On the other hand, the number of parameters in FMM
is larger than that in HCM. It is able to represent the differences
between categories more precisely than HCM, and thus is able to
resolve the two problems, described in Section 2, which plague HCM.

Another advantage of our method may be seen in contrast to the use of
{\em latent semantic analysis} \cite{Deerwester90} in document
classification and document retrieval. They claim that their method
can solve the following problems:
\begin{description} \item [synonymy problem] how to group synonyms,
  like `stroke' and `shot,' and make each relatively strongly
  indicative of a category even though some may individually appear in
  the category only very rarely; \item [polysemy problem] how to
  determine that a word like `ball' in a document refers to a `tennis
  ball' and not a `soccer ball,' so as to classify the document more
  accurately; \item [dependence problem] how to use dependent words,
  like `kick' and `goal,' to make their combined appearance in a
  document more indicative of a category.
\end{description} 
As seen in Tab.\ref{tab:softclus}, our method also helps resolve all
of these problems.

\section{Preliminary Experimental Results}

In this section, we describe the results of the experiments we have
conducted to compare the performance of our method with that of HCM
and others.

As a first data set, we used a subset of the Reuters newswire data
prepared by Lewis, called Reuters-21578 Distribution
1.0.\footnote{Reuters-21578 is available at\\ 
  http://www.research.att.com/\~lewis.} We selected nine overlapping
categories, i.e. in which a document may belong to several different
categories. We adopted the Lewis Split in the corpus to obtain the
training data and the test data.  Tabs.~\ref{tab:reuters1d} and
\ref{tab:reuters1c} give the details.  We did not conduct stemming, or
use stop words\footnote{`Stop words' refers to a predetermined list of
  words containing those which are considered not useful for document
  classification, such as articles and prepositions.}. We then applied
FMM, HCM, WBM , and a method based on cosine-similarity, which we
denote as COS\footnote{In this method, categories and documents to be
  classified are viewed as vectors of word frequencies, and the cosine
  value between the two vectors reflects similarity \cite{Salton83}.},
to conduct {\em binary} classification. In particular, we learn the
distribution for each category and that for its complement category
from the training data, and then determine whether or not to classify
into each category the documents in the test data.  When applying FMM,
we used our proposed method of creating clusters in Section 4 and set
$\gamma$ to be $0,0.4,0.5,0.7$, because these are representative
values. For HCM, we classified words in the same way as in FMM and set
$\gamma$ to be $0.5,0.7,0.9,0.95$.  (Notice that in HCM, $\gamma$
cannot be set less than $0.5$.)

\begin{table}[htb]
\begin{center}
  \caption{The first data set}
\label{tab:reuters1d}
\begin{tabular}{|lc|} \hline
Num. of doc. in training data & 707 \\
Num. of doc in test data & 228 \\
Num. of (type of) words & 10902 \\
Avg. num. of words per doc. & $310.6$ \\ \hline
\end{tabular}
\end{center}
\end{table}

\begin{table}[htb]
\begin{center}
  \caption{Categories in the first data set}
\label{tab:reuters1c}
\begin{tabular}{|l|} \hline
 wheat,corn,oilseed,sugar,coffee \\
 soybean,cocoa,rice,cotton \\ \hline
\end{tabular}
\end{center}
\end{table}

\begin{table}[htb]
\begin{center}
  \caption{The second data set}
\label{tab:reuters2d}
\begin{tabular}{|lc|} \hline
Num. of doc.  training data & 13625 \\
Num. of doc. in test data & 6188 \\
Num. of (type of) words & 50301 \\
Avg. num. of words per doc. & 181.3 \\ \hline
\end{tabular}
\end{center}
\end{table}

\begin{table}[htb]
\begin{center}
  \caption{Tested categories in the second data set}
\label{tab:reuters2c}
\begin{tabular}{|l|} \hline
 earn,acq,crude,money-fx,grain \\
 interest,trade,ship,wheat,corn \\ \hline
\end{tabular}
\end{center}
\end{table}

As a second data set, we used the entire Reuters-21578 data with the
Lewis Split. Tab.~\ref{tab:reuters2d} gives the details. Again, we did
not conduct stemming, or use stop words. We then applied FMM, HCM, WBM
, and COS to conduct {\em binary} classification.  When applying FMM,
we used our proposed method of creating clusters and set $\gamma$ to
be $0,0.4,0.5,0.7$.  For HCM, we classified words in the same way as
in FMM and set $\gamma$ to be $0.5,0.7,0.9,0.95$. We have not fully
completed these experiments, however, and here we only give the
results of classifying into the ten categories having the greatest
numbers of documents in the test data (see Tab.~\ref{tab:reuters2c}).

For both data sets, we evaluated each method in terms of {\em
  precision} and {\em recall} by means of the so-called
micro-averaging \footnote{In micro-averaging\cite{Lewis94b}, precision
  is defined as the percentage of classified documents in all
  categories which are correctly classified. Recall is defined as the
  percentage of the total documents in all categories which are
  correctly classified.}.  

When applying WBM, HCM, and FMM, rather than use the standard
likelihood ratio testing, we used the following heuristics. For
simplicity, suppose that there are only two categories $c_1$ and
$c_2$. Letting $\epsilon$ be a given number larger than or equal 0, we
assign a new document $d$ in the following way:
\begin{equation}\label{eq:testing}
\begin{array}{ll}
\frac{1}{N} (\log L(d|c_1) - \log L(d|c_2)) > \epsilon; &
d \to c_1, \\
\frac{1}{N} (\log L(d|c_2) - \log L(d|c_1)) \ge \epsilon; &
d \to c_2, \\
\mbox{otherwise}; & \mbox{unclassify $d$}, \\
\end{array}
\end{equation}
where $N$ is the size of document $d$. (One can easily extend the
method to cases with a greater number of categories.)
\footnote{Notice that words which are discarded in the clustering
  process should not to be counted in document size.} For COS, we
conducted classification in a similar way. 

\begin{figure}[htb]
\epsfxsize8.2cm\epsfysize6cm\epsfbox{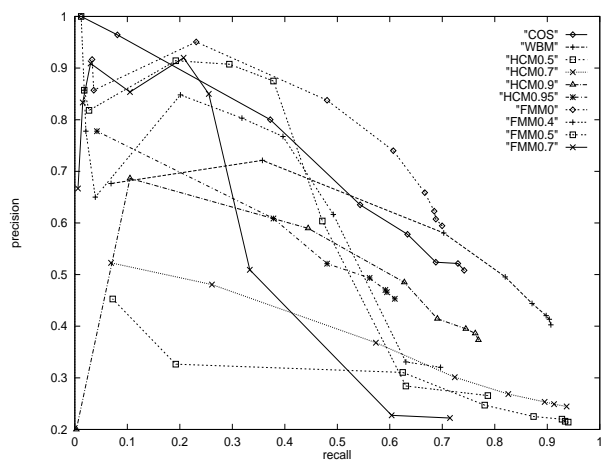}
\caption{Precision-recall curve for the first data set}
\label{fig:reuters1}
\end{figure}

\begin{figure}[htb]
\epsfxsize8.2cm\epsfysize6cm\epsfbox{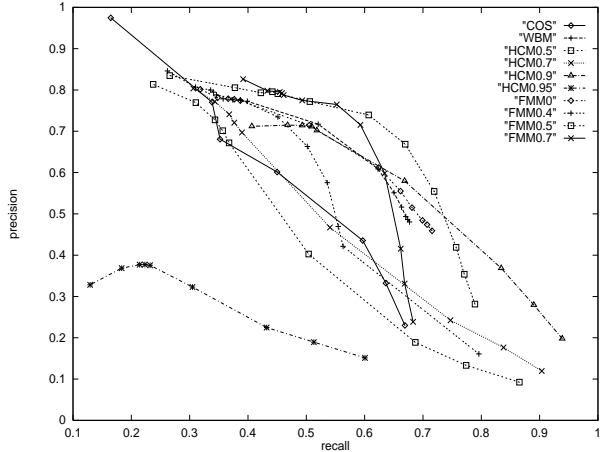}
\caption{Precision-recall curve for the second data set}
\label{fig:reuters2}
\end{figure}

Figs.~{\ref{fig:reuters1}} and {\ref{fig:reuters2}} show
precision-recall curves for the first data set and those for the
second data set, respectively. In these graphs, values given after FMM
and HCM represent $\gamma$ in our clustering method (e.g. FMM0.5,
HCM0.5,etc).  We adopted the break-even point as a single measure for
comparison, which is the one at which precision equals recall; a
higher score for the break-even point indicates better performance.
Tab.~\ref{tab:reuters1bp} shows the break-even point for each method
for the first data set and Tab.~\ref{tab:reuters2bp} shows that for
the second data set. For the first data set, FMM0 attains the highest
score at break-even point; for the second data set, FMM0.5 attains the
highest.

\begin{table}[htb]
\begin{center}
\caption{Break-even point for the first data set}
\label{tab:reuters1bp}
\begin{tabular}{|l|c|} \hline
COS & $0.60$ \\
WBM & $0.62$ \\
HCM0.5 & $0.32$ \\
HCM0.7 & $0.42$ \\
HCM0.9 & $0.54$ \\
HCM0.95 & $0.51$ \\
FMM0 & $0.66$ \\
FMM0.4 & $0.54$ \\
FMM0.5 & $0.52$ \\
FMM0.7 & $0.42$ \\ \hline
\end{tabular}
\end{center}
\end{table}

\begin{table}[htb]
\begin{center}
  \caption{Break-even point for the second data set}
\label{tab:reuters2bp}
\begin{tabular}{|l|c|} \hline
COS & $0.52$ \\
WBM & $0.62$ \\
HCM0.5 & $0.47$ \\
HCM0.7 & $0.51$ \\
HCM0.9 & $0.55$ \\
HCM0.95 & $0.31$ \\
FMM0 & $0.62$ \\
FMM0.4 & $0.54$ \\
FMM0.5 & $0.67$ \\
FMM0.7 & $0.62$ \\ \hline
\end{tabular}
\end{center}
\end{table}

We considered the following questions:

(1) The training data used in the experimentation may be considered
sparse. Will a word-clustering-based method (FMM) outperform a
word-based method (WBM) here?

(2) Is it better to conduct soft clustering (FMM) than
to do hard clustering (HCM)?

(3) With our current method of creating clusters, as the threshold
$\gamma$ approaches 0, FMM behaves much like WBM and it does not enjoy
the effects of clustering at all (the number of parameters is as large
as in WBM). This is because in this case (a) a word will be assigned
into all of the clusters, (b) the distribution of words in each
cluster will approach that in the corresponding category in WBM, and
(c) the likelihood value for each category will approach that in WBM
(recall case (2) in Section 3). Since creating clusters in an optimal
way is difficult, when clustering does not improve performance we can
at least make FMM perform as well as WBM by choosing $\gamma=0$. The
question now is "does FMM perform better than WBM when $\gamma$ is
$0$?"

In looking into these issues, we found the following:

(1) When $\gamma\gg 0$, i.e., when we conduct clustering, FMM does not
perform better than WBM for the first data set, but it performs better
than WBM for the second data set.

\begin{figure}[htb]
\epsfxsize8.2cm\epsfysize6cm\epsfbox{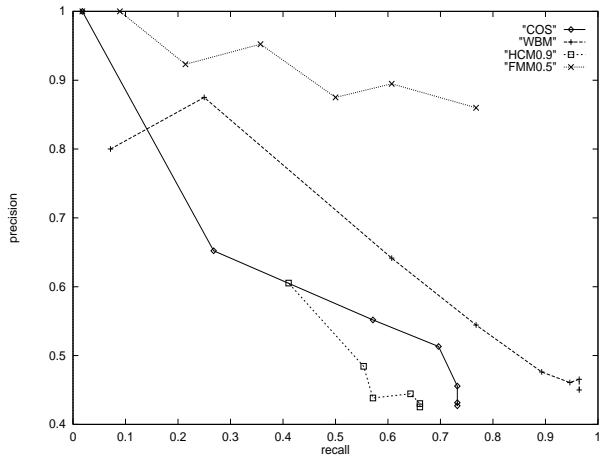}
\caption{Precision-recall curve for category `corn'}
\label{fig:corn}
\end{figure}

\begin{figure}[htb]
\epsfxsize8.2cm\epsfysize6cm\epsfbox{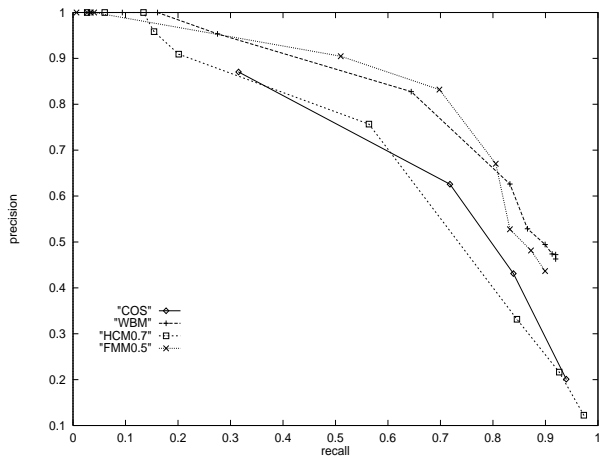}
\caption{Precision-recall curve for category `grain'}
\label{fig:grain}
\end{figure}

Evaluating classification results on the basis of each individual
category, we have found that for three of the nine categories in the
first data set, FMM0.5 performs best, and that in two of the ten
categories in the second data set FMM0.5 performs best. These results
indicate that clustering sometimes does improve classification results
{\em when we use our current way of creating clusters}.
(Fig.~\ref{fig:corn} shows the best result for each method for the
category `corn' in the first data set and Fig.~\ref{fig:grain} that for
`grain' in the second data set.)

(2) When $\gamma\gg 0$, i.e., when we conduct clustering, the best of
FMM almost always outperforms that of HCM.

(3) When $\gamma = 0$, FMM performs better than WBM for the first data
set, and that it performs as well as WBM for the second data set.

In summary, FMM always outperforms HCM; in some cases it performs
better than WBM; and in general it performs at least as well as WBM.

For both data sets, the best FMM results are superior to those of COS
throughout. This indicates that the probabilistic approach is more
suitable than the cosine approach for document classification based on
word distributions.

Although we have not completed our experiments on the entire Reuters
data set, we found that the results with FMM on the second data set are
almost as good as those obtained by the other approaches reported in
\cite{Lewis94b}. (The results are not directly comparable, because (a)
the results in \cite{Lewis94b} were obtained from an older version of
the Reuters data; and (b) they used stop words, but we did not.)

We have also conducted experiments on the Susanne corpus
data\footnote{The Susanne corpus, which has four non-overlapping
  categories, is available at ftp://ota.ox.ac.uk } and confirmed the
effectiveness of our method.  We omit an explanation of this work here
due to space limitations.

\section{Conclusions}

Let us conclude this paper with the following remarks:
\begin{enumerate}
\item The primary contribution of this research is that we have
  proposed the use of the finite mixture model in document
  classification.
\item Experimental results indicate that our method of using the
  finite mixture model outperforms the method based on hard clustering
  of words.
\item Experimental results also indicate that in some cases our method
  outperforms the word-based method {\em when we use our current
    method of creating clusters}.
\end{enumerate}

Our future work is to include:
\begin{enumerate}
\item comparing the various methods over the entire Reuters corpus and
  over other data bases,
\item developing better ways of creating clusters.
\end{enumerate}

Our proposed method is not limited to document classification; it can
also be applied to other natural language processing tasks, like word
sense disambiguation, in which we can view the context surrounding a
ambiguous target word as a document and the word-senses to be resolved
as categories.

\section*{Acknowledgements}

We are grateful to Tomoyuki Fujita of NEC for his constant
encouragement. We also thank Naoki Abe of NEC for his important
suggestions, and Mark Petersen of Meiji Univ. for his help with the
English of this text. We would like to express special appreciation to
the six ACL anonymous reviewers who have provided many valuable
comments and criticisms.

\end{document}